# Social influence under uncertainty in interaction with peers, robots and computers

Joshua Zonca*[1], Anna Folsø [2], Alessandra Sciutti [1]

1. Cognitive Architecture for Collaborative Technologies (CONTACT) Unit, Italian Institute of Technology, Genoa, Italy.
2. Department of Informatics, Bioengineering, Robotics and Systems Engineering, University of Genoa, Genoa, Italy.

**Corresponding author:** Correspondence to Joshua Zonca, Italian Institute of Technology, Via Enrico Melen, 83, 16152 Genova, GE (Italy). Email: joshua.zonca@iit.it

**Acknowledgements:** We gratefully acknowledge the financial support of the European Research Council (ERC Starting Grant 804388, wHiSPER).

**Competing interests:** The authors declare no competing interests.

## Abstract

Taking advice from others requires confidence in their competence. This is important for interaction with peers, but also for collaboration with social robots and artificial agents. Nonetheless, we do not always have access to information about others' competence or performance. In these uncertain environments, do our prior beliefs about the nature and the competence of our interacting partners modulate our willingness to rely on their judgments?

In a joint perceptual decision making task, participants made perceptual judgments and observed the simulated estimates of either a human participant, a social humanoid robot or a computer. Then they could modify their estimates based on this feedback. Results show participants' belief about the nature of their partner biased their compliance with its judgments: participants were more influenced by the social robot than human and computer partners. This difference emerged strongly at the very beginning of the task and decreased with repeated exposure to empirical feedback on the partner's responses, disclosing the role of prior beliefs in social influence under uncertainty. Furthermore, the results of our functional task suggest an important difference between human-human and human-robot interaction in the absence of overt socially relevant signal from the partner: the former is modulated by social normative mechanisms, whereas the latter is guided by purely informational mechanisms linked to the perceived competence of the partner.

## Keywords

# 1. Introduction

In our everyday life, we are often influenced by our peers. This process is not blind: we need to choose *when* relying on others and *whom* we should trust [1, 2]. However, this is not a straightforward task: indeed, we often implement suboptimal strategies in the processing and use of information provided by others. In this regard, extensive research on social cognition and decision making has shown that, in social contexts, human judgment is distorted by peculiar cognitive biases (e.g., [3-5]) and social normative mechanisms (e.g., [6-8]).

Nowadays, the theme of social influence is not only important in the context of interaction with peers, but also in interaction with other types of autonomous agents, including social robots. Research in fields such as Human-Robot Interaction (HRI) and social robotics aims at developing robots that could assist humans in their activities [9-12]. Robots are already utilized to deliver a vast range of services to workers, patients and customers. For instance, they provide information to guests and customers in hotels, restaurant and stores [13-15], assist older people at home [16, 17], support diabetes self-management in children [18], provide financial advice [19] and participate in education programs [20-21]. In all these contexts, robots should be designed with the goal of leading humans to accept help from robots and follow their advice in collaborative scenarios. In this respect, research in HRI has shown that the main factor determining reliance on robots is its performance [22-26]. Precise feedback on robots' performance in the current task is fundamental to drive the formation of well-informed beliefs on their functional capabilities and reliability. On the contrary, in the presence of uncertainty about robots' competence, humans may over-trust or under-trust robots, resulting in suboptimal or even harmful behavior [27-31]. This is not a minor issue, since we can think of several real-life contexts in which human-robot interactions do (or might) not produce factual, exogenous feedback about robots' performance. For instance, robots may be used to convince customers to invest their money in a certain way, or apply healthy behavioral practices. In these circumstances, individuals should decide whether to *trust* their robotic informants before obtaining a factual feedback on the result of their actions (e.g., the investment has been profitable; the advice has improved my health). In these contexts, it might be extremely difficult to determine the reliability of our interacting partner and choose whether to follow or not the robot's advice. This task may be more difficult than in interaction with human peers (e.g., financial advisors, doctors). Indeed, humans have rather sophisticated beliefs on the competence and the ability of their peers in a vast series of task, but their "a priori" knowledge of robots and their skills is much less precise. In this sense, our prior beliefs about the relational and informational characteristics of our robotic partner may distort our process of weighting of information by its reliability, leading to over- or under-use of the robot itself [32-33].

The present study aims at investigating how humans perceive and use information on other agents' behavior in the presence of uncertainty about their performance. In particular, our goal is to understand whether and how our prior beliefs on human peers and social robots may distort our estimates of their performance and, consequently, our reliance on them. In this regard, no study to our knowledge compared *directly* our willingness to follow the opinions of human peers and social robots in judgment under uncertainty. Indeed, it would be extremely important to analyze individuals' behavior during interaction with agents of different natures (e.g., human or robot) that produce *identical* functional performances in the same social influence scenario. This would allow us to test directly the role of prior beliefs and a priori biases about human and robotic agents in the interpretation of their functional and social normative properties. Importantly, our study will provide novel insights on how robots, when compared to humans, can be perceived and trusted in the absence of precise feedback about their reliability and performance. These findings will shed new light on the mechanisms driving the acceptance of information delivered by robots, which will be useful in several real-life contexts in which robots should help, assist and inform humans.

## 2. Related work

### 2.1. Social influence in human-human interaction.

Humans take advice from other human beings to obtain novel knowledge and optimize their decisions. This form of social influence reduces behavioral uncertainty [2, 34] and the costs and the risks associated with individual learning based on trial and error [1, 35, 36]. Extensive research has shown that humans use information provided by peers when they evaluate it as reliable, whereas they discard it otherwise [37-40]. In order to evaluate the reliability of information provided by others and choose the best sources of advice, humans extract information about others' confidence [41, 42] and expertise [43-46]. Nonetheless, in many situations humans use suboptimal informational criteria in the process of evaluation and use of information acquired through the social channel.

One the one hand, suboptimality can emerge due to cognitive biases: for instance, people tend to over-discount socially acquired information [47-49], over-estimate their own abilities [50, 51] and under-estimate own responsibility in the case of negative outcomes in collaborative scenarios [5, 52]. Moreover, people assign an excessive weight to opinions confirming their own original beliefs and ignore contradictory ones in the presence of different information sources [53]. These egocentric distortions lead to a relatively low (and often suboptimal) reliance on others' opinions, even if they are more competent than us [54]. Importantly, these biases seem to be inherently linked to the *social* nature of the information source: indeed, the same piece of information is trusted more if acquired though individual exploration than social transmission [3, 55].

On the other hand, distortions in the integration of information provided by others can be linked to relational and social normative dynamics. In social environments, humans are willing to sacrifice accuracy to modulate reputation, social status and affective rewards. For instance, humans are inclined to conform to the opinion of their peers in order to control their reputation in the group and affiliate with them [6, 7, 56-58]. Moreover, the individual willingness to change idea in favor of that of others is modulated by reciprocity in adult dyads [8, 59] and triads [60], and also in dyadic child-adult interactions [61]. In other words, we are more prone to change our ideas in favor of others if they have endorsed our opinions in the past.

These findings highlight the emergence of peculiar mechanisms underlying sharing and use of information in human-human interaction, which are modulated by cognitive distortions and social norms.

## 2.2. Social influence in human-robot interaction

Previous research in HRI has investigated when and how humans accept help and advice from a robot. In this respect, we highlight that, in the HRI literature, the concept of social influence often overlaps with the concept of "trust": in this sense, if a human being "trusts" a robot is willing to rely on it (for a discussion on the different conceptualizations of trust in HRI, see [62]). In most of the cases, researchers have been focusing on investigating what robots' characteristics affect humans' reliance on robots. Numerous studies have shown that the main factor modulating reliance on robots is its performance [22-26]. Humans tend to "trust" robots as long as they show reliable behavior, but they rapidly lose confidence in its competence in in presence of failures [63-65], potentially leading to disuse of the robotic system [32-33]. However, people sometimes trust robots more than they should. For instance, several studies have shown that people tend to over-comply with the instructions of robots, even if they have previously show faulty or unreliable behavior [27-31].

In the last years, research in HRI have been focusing on "social" features of robots. The main idea is that the introduction of a *humane* dimension in robots (i.e., *social* robots) can be beneficial for human-robot interaction [10, 11]. In line with this approach, extensive evidence has shown the emergence of pro-social attitudes towards social robots [66-69] and improvements in human-robot collaboration [70-73]. In some cases, robots with human-like traits can increase humans' trust towards them [74]. For instance, robots showing empathic behavior can be more persuasive [75] and convince people to reduce unhealthy behavior (e.g., alcohol consumption, [76]).

Nevertheless, evidence on the emergence of relational and social normative mechanisms modulating human-robot interaction is rather limited and context-dependent. Compliance towards robots tends to emerge primarily on functional tasks (i.e., tasks in which the robot has to fulfill a concrete goal through actions or quantitative judgment) rather than social ones (i.e., judgments or decisions on social issues), as shown in recent studies [77, 78]. Some experiments tried to replicate classical effects of normative conformity with mixed results [79-84]. In particular, the effect of social norms and relational dynamics on social influence seems to be unable to overcome the functional effects linked to uncertainty and performance. In this sense, compliance is intrinsically tied with the difficulty of the task solution: individuals typically conform to robots only if they are unsure about the action to take [85] and tend to resist their influence otherwise.

# 3. Present study

Independent streams of research on human-human and human-robot interaction have been investigating whether and how individuals choose to rely on peers or robots to optimize their decisions. Most of the studies (especially in HRI) employ factual feedback about the robot's performance, which, not surprisingly, has been shown to be the most decisive factor explaining reliance on robots. However, many of the real-life scenarios using robots to provide services to customers and patients require humans to take decisions (i.e., follow or not a robot's advice) well before obtaining feedback on the relative outcomes. In these uncertain contexts, it is important to understand how prior beliefs on robots shape human decisions in human-robot social influence settings and how these decisions compare to those taken in the presence of human advice.

The present study aims at designing an experimental protocol that allows a direct comparison between decisions taken during interaction with different types of partner (human peer, humanoid robot, computer algorithm) in the presence of uncertainty about the interacting agents' performance. To achieve this goal, in the current work we gather data from two recent studies (originally) investigating reciprocity of social influence in two different interactive contexts: human-human [60] and human-robot [79] interaction. For the purposes of the current study, we will focus on a joint perceptual task (Social influence task) that has been used in the above-mentioned studies to assess participants' willingness to follow the advice of a partner. To date, the task has been used only in independent interactive contexts (human-human or human-robot interaction) and only as baseline measure in conjunction with other joint tasks, in order to assess the role of reciprocity in social influence. Instead, the current study aims at directly comparing social influence across different interactive contexts (human-human, human-robot and also human-computer interaction) to investigate the role of prior beliefs on the nature of our interacting partners in determining our susceptibility to their judgments. Nevertheless, we acknowledge that a set of results reported in [60] and [79] have been used in the current manuscript (See the paragraph "6.2.1. Estimation error: perceived vs. actual performance" in the Results section). We believe that reporting these findings in the current manuscript will help the reader in the understanding of important aspects of participants' behavior that are relevant for the interpretation of the following novel results on the role of prior beliefs in social influence under uncertainty.

The advantage of investigating social influence mechanisms using a joint perceptual task lies in the opportunity to characterize participants' and partners' behavior through extremely controlled and precise responses, which could be transparently and unambiguously tracked by the interacting partners along the task though online feedback. The use of perceptual tasks to investigate social influence processes goes back to a long tradition of research in psychology and cognitive science starting with the classical Asch conformity experiment [6]. In our experimental paradigm, participants believed to interact with one of three different partners: a human participant, a humanoid social robot iCub [40, 41] or a computer. The last scenario represents a control condition in which the interacting agent does not have a social value and does not possess a perceptual system, but produces the same judgments. Participants performed a perceptual inference task by estimating the length of visual stimuli without any feedback and received trial-by-trial feedback revealing the judgment made by their partner. The partner's estimates in all the conditions were generated by the same algorithm, which produced rather accurate (but not perfect) responses. Then, in each trial, after observing the partner's response, participants could modify their own judgment by selecting any position between their own and their partner's estimate (final decision). The participants' shift from their own original estimate towards that of the partner in their final decision has been used as an index of influence from the partner' response. To guarantee that the manipulation modulated only prior beliefs about the interacting partner, we did not provide any feedback about the participant's or the partner's accuracy.

## 4. Hypotheses

H1: Prior beliefs play a crucial role in the interpretation and use of feedback coming from an interacting partner. Indeed, we predict that the exact same response produced by different partners (human, robot or computer) will lead participants to *trust* and therefore *use* such feedback to different extents. Specifically, we hypothesize that the effect of prior beliefs will lead to a higher level of susceptibility to the advices of the robotic partner compared to human and computer ones, given the high "a priori" confidence that humans generally attribute to robots [27, 65, 74] and growing evidence highlighting the emergence of over-trust towards robots [28-31]. Moreover, we expect this effect to emerge strongly at the very beginning of the task, before exposure to repetitive feedback on the partner's behavior, when prior beliefs should have a higher impact on the interpretation of feedback on the partner's estimates. In this regard, we also predict that the interaction between prior beliefs and early factual feedback on the partner's behavior modulate participants' willingness to use the partner's advice in the following interactions.

H2: Interaction with human peers is underlined by peculiar relational and social normative dynamics, which are much less pervasive during interaction with non-human agents (humanoid social robots and computers). Specifically, we expect that in the Human condition participants' susceptibility to the partners opinion may be explained by a lower degree by the perceived reliability of their partner's responses. In contrast, we predict that participants' willingness to follow the opinion of their partner in Robot and Computer conditions will be explained largely by the participants' estimated competence of the partner, without distortions arising from normative principles.

## 5. Materials and methods

### 5.1. Participants and procedure

#### 5.1.1. Overview

We analyze data of 75 participants (42 females, mean age: 32.96, SD: 12.30). All participants completed two different tasks in this exact order: Perceptual inference task and Social influence task. Participants belong to three between-subject groups of 25 participants each: Human, Robot and Computer. In all the conditions, participants performed the Perceptual inference task alone and then they performed the Social influence task with a partner. The difference between the three conditions lied in the participants' belief about the nature of the interacting partner: a human participant (Human condition), a humanoid robot iCub (Robot condition), or a computer (Computer condition). However, the partner's choices in all conditions was controlled by the same algorithm. All the participants performed the task using a touch screen tablet through which the participant and the (alleged) partner (human, robot or computer) could 1) make choices and 2) receive online feedback on the choices of the partner. Participants in all the conditions, including the Robot one, could not see their partner while performing the tasks. We chose not to design a task consisting in an on-site physical interaction between participants and their partner to guarantee the comparability of our experimental conditions from a functional perspective. For instance, participants in the Human condition would likely be biased by the individual characteristics of the partner. Moreover, in the Robot condition, we did not want that participants could infer the robot's ability in the task by observing the characteristics and kinematics of its movements. In all the conditions, we wanted participants to acquire knowledge about their partner's ability just by observing its responses, in order to understand the role of prior beliefs about the nature of the interacting agent in the perception and use of its functional feedback. Nonetheless, participants in the Robot condition, before starting the experimental tasks, had the possibility to meet their partner in the experiment (a humanoid robot iCub). This was important since all the participants in the Robot conditions were naïve about robots and their functioning. In this way, they could get an idea of the type of agent they would have interacted with, although they did not receive any information about the factual accuracy of the robot in its perceptual estimates, to allow the comparability with the other two conditions. For a complete description of the participants' encounter with the robot, see the next paragraph "Introducing the humanoid robot iCub".

Participants in the Human and Computer conditions underwent the same experimental paradigm as the ones assigned to the Robot condition, but they did not meet the robot before the start of the experimental protocol. In the Human condition, participants were told that their partner had been recruited with the same modalities and was performing the task in a neighboring experimental room. Participants in the Human condition were explicitly told that this aspect of the experimental procedure was necessary to prevent participants from knowing the actual characteristics of their human partner, which could bias their behavior. Participants in the Computer condition were informed that the partner's responses were generated by a computer, without any additional information on how these responses would be generated.

For all participants, the experimental paradigm was carried out in a dimly lit room, to ensure an optimal visibility of the stimuli on the screen. Participants were seated in front of a wide touch screen (43.69 x 24.07 cm), at a distance that allowed participants to see the visual stimuli and reach the screen with a touch-pen with an ultrathin tip, which allowed participants to respond with high spatial accuracy. Before beginning the experimental tasks, written instructions were given and participants were allowed to ask questions.

Participants were told that their reimbursement would have been calculated based on their performance and, in particular, on the accuracy of both their initial and final estimates in both tasks. The accuracy of the partner was not supposed to have an impact on participants' outcomes, and vice versa. Anyway, everyone received a fixed amount at the end of the experiment, following the guidelines of the Italian Institute of Technology and the local ethics committee (Azienda Sanitaria Locale Genovese N.3, protocol: IIT_wHiSPER) concerning the application of a fair reimbursement for voluntary participation in experimental research.

At the end of the experiment, participants were asked indirect questions about the experiment and the hypothesized characteristics and abilities of their partner, in order to assess whether they have believed the cover story and the reimbursement procedure. All participants have shown to believe the cover story and that their final reimbursement would be affected by their performance. Eventually, we extensively debriefed participants about the experimental procedures, the reasons underlying the modality of the experiment, the reimbursement procedure and the goals of our research, in accordance with the relevant ethical guidelines.

### 5.1.2. Introducing the humanoid robot iCub

Before starting the experiment, participants in the Robot condition, had a brief meeting with iCub [86, 87], which acted as their partner in the Social influence task. iCub is an open source humanoid robot designed for research in embodied cognition and artificial intelligence. It is characterized by 53 actuated degrees of freedom that allow fine-grained movements of head, arms, hands, waist and legs. It is endowed with sensors and actuators enabling the generation of controlled and precise actions and the direction of its gaze towards objects and individuals. The robot possesses lines of LEDs representing mouth and eyebrows mounted behind the face panel to produce facial expressions and simulate speech. All these characteristics allow iCub to show human-like behavior and appearance [86, 88] and to be perceived as an intentional agent [89-93], capable of generating autonomous actions to accomplish specific goals. In the current study, the robot should appear to participants as 1) a social and intentional agent, which was aware of the presence of the participant and knew about the upcoming joint experiment and 2) an embodied agent that could physically perform the same task that participants would face in their experimental session.

For these reasons, participants in the Robot group met iCub before starting the experimental tasks. Before the participants' arrival, iCub was placed in front of a touch-screen tablet, which was identical to the one that participants would use during their experimental session. Once the participant had arrived at the research center, one experimenter accompanied them in a dedicated room with iCub, while another experimenter controlled the robot from the sidelines. The robot made a series of predetermined actions (identical for all participants) by a custom-made script running in the YARP environment [94]. The researcher controlling the robot managed the timing of the robot's actions in order to simulate a natural interaction with the participant. Once the participant had entered in the room, iCub turned towards them and said hello with both its voice and hand. The participant was accompanied in front of the robot, in order to allow iCub to track their face and direct its head and gaze towards them. The robot introduced itself and told the participant that they would play a game together; meanwhile, iCub continued to look at the participant and follow their head movements. Through these actions, we aimed at signaling to the participant that iCub was aware of their presence and knew that it would interact with them in the upcoming experiment. Then iCub said goodbye to the participant, turned towards the tablet and announced that it was ready to start the experiment. To give the impression that iCub could observe stimuli on the touch-screen tablet and interact with it to perform the task, the robot also leaned forward and directed its right arm and hand towards the tablet, pointing in the direction of the screen with its right index, as if it was ready to touch it. Eventually, the participant was guided in another room to start the experiment. We highlight that participants did not receive any direct or indirect information about the robot's accuracy in the task (see the paragraph "5.2.3. Agents' behavior in the Social influence task"), in order to guarantee the comparability of the three experimental conditions.

## 5.2. Tasks description

### 5.2.1. Perceptual inference task

The perceptual inference task was identical for all participants in the three experimental conditions. In each trial (Fig. 1), two red disks (diameter = 0.57 cm) appeared one after the other on a visible horizontal white line. The first disk appeared at a variable distance from the left border of the screen (0.6–6.6 cm) and remained on screen for 200 ms before disappearing. After an inter-stimulus interval of 200 ms, a second disk appeared at a variable distance to the right of the first disk. This distance has been defined as the target stimulus length (s) and was randomly selected from 11 different sample lengths (min: 8 cm, max: 16 cm, step: 0.8 cm). The second disk disappeared after 200 ms, as the first disk. Then participants had to touch a point on the horizontal line, to the right of the second disk, in order to reproduce a segment (connecting the second and the third disk) that matched the target stimulus length. Right after the participant's screen touch, a third red disk appeared in the selected location. We did not provide any feedback about the accuracy of the response. The task consisted of 66 trials. At the end of the task, participants were asked to evaluate from 1 to 10 their accuracy in perceptual estimation.

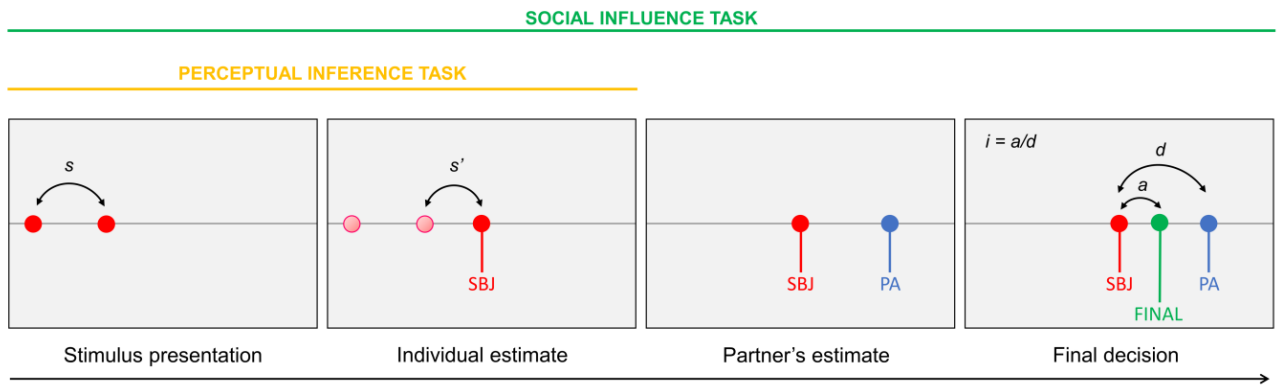

**Fig. 1.** Perceptual inference task and Social influence task. Perceptual inference task. Participants were presented with a red disk that appeared on a horizontal line and then disappeared after 200 ms, followed by another disk appearing and disappearing after 200 ms to the right of the first disk. Participants were asked to touch a point (SBJ), to the right of the second disk, in order to reproduce the stimulus length (s'), namely the distance between the first and the second disk (s). Social influence task. In each trial, first the participant had to estimate the length of a visual stimulus, as in the Perceptual inference task and then received feedback on the estimate of their partner (PA). Eventually, the participant made a final decision by choosing a point between own and partner's estimates (FINAL). The index of influence (i) was computed as the adjustment towards the partner in the final decision (a) divided by the distance between the two agents' responses (d).

**5.2.2. Social influence task**

In each trial, participants performed length estimation of a visual stimulus as in the Perceptual inference task. Participants were told that, during this interval, the very same stimuli would have been presented also to their interacting partner (computer in the Computer condition, iCub in the Robot condition, human participant in the Human condition), that would have made its own perceptual estimate. Right after their estimate, a vertical red line with the word YOU appeared at the exact response location. After the participants' estimate, the other agent's estimate was shown, marked with the word PC (Computer condition), ICUB (Robot condition) or BP (Blue Player, Human condition) in blue (Fig. 1). Moreover, participants were told that also their response would be shown to their interacting partner. Importantly, the partner's response was always shown *after* the participant's estimate, with a jittered delay, so that this information could not be used by participants in their perceptual estimation process. The partner's feedback was shown 1.50-2.75 seconds after the participant's response. The magnitude of the delay was increased in the Robot and Computer conditions (2.50-2-75 s), compared to the Human condition (1.50-1.75), to be more credible: indeed, a robot requires some seconds to make a spatially-precise response on a delicate surface like a touch-screen tablet, as also visible in the meeting with the robot before the experiment (see the paragraph "Introducing the humanoid robot iCub"). The delay in the Computer condition was set to match that of the Robot condition. The delay in the Human condition was shorter since participants would likely assume similar reaction times for another human participant, whereas long delays would have casted doubts for the credibility of the cover story.

Anyway, it is very unlikely that the delay in the partner's response had an impact on the perceived competence of the agent, since participants in the different conditions were informed about the presence of an arbitrary and variable delay imposed by the experimenters. Indeed, participants were explicitly told, before starting the experiment, that the timing of the partner's response feedback on the participant's touch-screen tablet was *postponed* by a variable delay starting from the actual reaction time of the partner. The final debriefing confirmed that none of the participants believed that the timing of appearance of the partner's response revealed significant information for the assessment of the partner's accuracy.

After the counterpart's estimate, participants had to make a final decision: they could select any position between their previous response and the partner's response, but not outside this range. In these terms, participants' final decisions express the relative weight assigned to the judgment of the two interacting partners. After the participant's final decision, a green dot and a vertical green line with the word FINAL marked the position chosen by the participant. Participants were told that information about both their initial estimate and final decision would have been shown to their partner. The task consisted of 66 trials divided in three blocks by two pauses. The position of all the three responses (participant's estimate, partner's estimate and final decision) remained visible on the screen for 1 s.

At the end of the task, participants were asked to evaluate from 1 to 10 their own and the other agent's accuracy in in the perceptual estimates, without taking into account their final decisions.

### 5.2.3 Agents' behavior in the Social influence task

The simulated perceptual judgments of the participants' partner (human, robot or computer) in the Social influence task were controlled by the same algorithm. In each trial, the actual length of the agent's perceptual estimate was randomly drawn from a gaussian distribution centered at the current stimulus length (i.e., the correct response). The response distribution was characterized by a standard deviation of 1.52 cm. The magnitude of the standard deviation of the response distribution was chosen to maintain a balance between accuracy, variability and credibility of the agent's behavior. We aimed at controlling the variance of the algorithm in order to prevent participants from recurrently observing extremely high discrepancies between the two responses, which would have affected considerably the perceived reliability of the partner. For this reason, the standard deviation of the response distribution of the agent was set to be 25 % lower than the observed standard deviation of participants' responses, as estimated in a pilot study. However, we also considered the possibility that few participants could be very accurate in their perceptual inferences. In this case, the participant and their partner would select close responses very often, preventing the emergence of variability in participants' final decisions. Therefore, in half of the cases in which the sampled estimate of the algorithm was rather close to the participant's one (i.e., d < 0.5 cm), the algorithm re-sampled a new response from the distribution (i.e., until d > 0.5 cm).[1]

This response distribution was used for all the perceptual estimates in all the three between-subject conditions of the Social influence task.

### 5.3. Data analysis

### 5.3.1. Perceptual inference task

For the purpose of subsequent between-condition analyses in the Social influence task, we compared the performance of the three groups (Computer, Robot and Human) to ensure that they were comparable in terms of perceptual abilities in the Perceptual inference task. The parameter of accuracy that has been used throughout the paper was the *estimation error*, computed as the absolute difference between the estimated length and the actual stimulus length (e), divided by stimulus length (s):

$$estimation\ error = \frac{e}{s}$$

---

[1] The value of 0.5 cm has been chosen after a pilot study aimed at assessing participants' perception of agents' distance (i.e., discrepancy between participant's and partner's estimate). Values around 0.5 cm (or lower) were generally interpreted as "minor" distances. Nonetheless, we resampled agent's estimates in the case of distance lower than 0.5 cm only in half of the cases, since the full absence of trials with minimum discrepancy between the two responses would have appeared as suspicious (e.g., participants might have thought that the partner's responses were specifically set to be far from the participants' responses). Anyway, we highlight that distances lower than 0.5 cm were infrequent in the Social influence task (7% of the total number of trials).

Dividing the actual estimation error by the current stimulus length allows us to give equal weights to trials including short and long visual stimuli.

### 5.3.2. Social influence task

In the Social influence task, participants' made their perceptual estimates as in the previous task and then could observe the partner's estimate concerning the same stimulus. Then they were asked to make a final decision, placing their final response in any position between their own and the partner's response. Concerning participants' final decisions, an index of *influence* was computed as the discrepancy between the participants' final and initial response (*a*) divided by the distance between the two agents' initial estimates (d):

$$influence = \frac{a}{d}$$

Therefore, the index of influence can take *any value* between 0 and 1, where, for instance, 0 indicates a final decision coinciding with the participant's own estimate, 0.5 reflects an equal distribution of weight towards own and other's judgments and 1 corresponds to a final decision coinciding with the partner's estimate.[2]

Influence was compared across experimental conditions (Human, Robot, Computer) to investigate the effect of prior beliefs on the processing of the other agent's feedbacks. To further investigate the hypothesis that prior beliefs about the partner's capabilities are crucial for determining the participants' willingness to follow the partner's responses in their final decisions, we analyzed the evolution of *influence* over time and across conditions. In particular, we hypothesize that differences in participants' influence across conditions are relatively higher at the beginning of the experiment, when prior beliefs are likely to exert the highest impact on the interpretation of available feedback.

We also tested whether participants used available feedback revealing (indirect) information about participants' own and partner's competence to decide whether to use their partner's estimate in their final decisions. We defined the index of *agents' distance* as the absolute distance between the participant and the partner's estimates (d) divided by the current stimulus length (s):

$$agents'\ distance = \frac{d}{s}$$

---

[2] In a few cases, along the manuscript, the index of influence will be expressed as a percentage instead of a proportion (e.g., influence = 0.14 or influence =14%). For instance, an influence of 100% will express full reliance in the partner's estimates and an influence of 50% will express an equal distribution of weight towards own and other's judgments.

As for the estimation error, by weighting distance by the actual stimulus length we can assign equal weights to short and long stimuli. The agents' distance is indeed the only feedback available to participants to estimate their own and their partner's competence in the Social influence task. Therefore, we explored the relationship between trial-by-trial agents' distance and influence to investigate the impact of the discrepancy between the responses of the two interacting agents on the participants' willingness to follow the partner's opinion.

We also investigated whether participants' influence along the experiment could be explained by the interaction between prior beliefs about their partner's competence and the first impression on the reliability of the partner's responses, as expressed by the agents' distance in the first trial of the experiment. This hypothesis is in line with the idea that prior beliefs, which are supposed to play a crucial role in the interpretation of the partner's responses, should have a strong impact in the interpretation of *early* feedback linked to others' competence. Conversely, the exposure to a relatively long history of feedback on the partner's responses should mitigate the effect of prior beliefs, leading to higher reliance on processes of estimation of the partner's abilities based on empirical feedback. To test this hypothesis, we considered the agents' distance in the very first trial of the Social influence task and, for each of our three samples, we performed a median split to divide participants that observed a low distance from participants that observed a high distance from the partner at the beginning of the experiment. We will refer to this categorical factor as *initial distance* (low or high). Then we compared participants' influence across the two levels of the initial distance factor, in the three different conditions, to assess the impact of the very first factual feedback on participants' perception of the partner competence.

Eventually, we tested whether and how participants' susceptibility to the partner's judgments in the three conditions was modulated by the perceived reliability of the partner's estimates. We analyzed participant's performance ratings concerning own and partner's accuracy (from 1 to 10), which were collected at the end of the Social influence task. In particular, we tested the relationship between performance ratings (i.e., self – other ratings) and influence to assess whether participants' willingness to follow the partner's opinion was well-explained by the perceived competence of the partner.

### 5.3.3. Statistical data analysis

Most of the analyses reported in this work focus on analysing variation of task-related dependent variables (i.e., influence, estimation error) as a function exogenous experimental factors (i.e., experimental conditions) and endogenous predictors (i.e., agents' distance, performance ratings). Moreover, throughout the paper we also directly compared individual variables (e.g., mean influence, mean estimation error, performance ratings) across experimental conditions. Since these individual variables occasionally show some degree of skewness and, in some conditions, show a violation of the normality distribution assumption, we used non-parametric tests (Wilcoxon signed-rank test, Wilcoxon rank-sum test, Kruskal-Wallis test) throughout the paper for consistency. All tests are two-tailed and report z statistic, p-value and effect sizes ($r$, $\eta^2$). The formulas used for the calculation of the effect sizes can be found in [95] and [96]: r = Z/√N; $\eta^2$ = $Z^2$/N, where N is the total number of observations. For the same reason, we used non-parametric correlation tests (Spearman's rank correlation). Multiple comparisons have been treated using Bonferroni correction. The alpha level was set at p = 0.05 for all the statistical analyses.

Regression analysis was also used to explore relationships between our variables of interest (e.g., influence) and task-relevant predictors. Simple and multiple linear regressions were used to assess between-subject relationships between participants' individual measures (e.g., participants' performance ratings, participants' mean influence). Mixed-effects models were used to test relationships between within-subject levels of relevant predictors as well as trial-by-trial measures of interest (e.g., trial-by-trial influence, estimation error and agents' distance). In mixed-effects models, the random effect has been applied to the intercept to adjust for the baseline level of influence of each subject and model intra-subject correlation of repeated measurements. In regression and mixed-effects model analyses, we report unstandardized (B) and standardized (β) regression coefficients, t (or z) statistics and p-values. In one of the models (model 3), we have also treated the influence variable as a percentage (instead of a proportion), in order to facilitate the interpretation of the results. In this case, we report regression coefficients as B%, in addition to the classical unstandardized B values. In the Supplementary information, we report complete results of complex models (i.e., mixed-effects models), including tables with standard errors and confidence intervals. Specification and results of all the mixed-effects models have been described in detail in the Supplementary information.

All analyses include the entire sample of 75 subjects and all trials of the two experimental tasks.

### 5.3.4. Data availability

The datasets supporting analyses and figures included in the current study are available in a dedicated OSF repository at: https://osf.io/yk2cv/?view_only=40765e2bad2145a4834e6fad60699bd0

## 6. Results

### 6.1. Perceptual inference task

We investigated the comparability of our three experimental groups by testing for the presence of a difference in estimation error across experimental conditions. Results show that the three groups did not differ in terms of estimation error (Computer: 0.20 ± 0.05; Robot: 0.19 ± 0.06; Human: 0.21 ± 0.05. Kruskal-Wallis test, $\chi^2$ = 3.10, p = 0.212). Therefore, the null hypothesis assuming that the estimation error distribution is the same across the three groups cannot be rejected.

### 6.2. Social influence task

#### 6.2.1. Estimation error: perceived vs. actual performance

First, analysis of participants' estimation error reveals participants' accuracy in perceptual estimation was markedly lower than that of their partner in all the three experimental conditions (Wilcoxon rank-sum test, estimation error as dependent variable, agent (participant or partner) as independent variable. Computer: $z = 6.06$, $r = 0.86$, $\eta^2 = 0.73$, $p < 0.001$; Robot group: $z = 6.06$, $r = 0.86$, $\eta^2 = 0.73$, $p < 0.001$. Human: $z = 5.64$, $r = 0.80$, $\eta^2 = 0.64$, $p < 0.001$. Results are significant at the Bonferroni-corrected threshold for 3 comparisons). Moreover, in all conditions, participants' estimation error was comparable to that shown in the Perceptual inference task, suggesting that the partner's feedback did not help participants in improving their performance (Wilcoxon signed-rank test, estimation error as dependent variable, task as independent variable. Computer: $z = 0.040$, $r = 0.01$, $\eta^2 = 0.00$, $p = 0.968$; Robot: $z = -0.740$, $r = 0.10$, $\eta^2 = 0.01$, $p = 0.459$; Human: $z = 0.417$, $r = 0.06$, $\eta^2 = 0.00$, $p = 0.677$).[3]

The partner's average estimation error was comparable across conditions (Computer: $0.10 \pm 0.01$; Robot: $0.10 \pm 0.01$; Human: $0.10 \pm 0.01$. Kruskal-Wallis test, $\chi^2(2) = 2.29$, $p = 0.318$). As in the Perceptual inference task, we did not find any difference in terms of participants' estimation error across conditions (Computer: $0.21 \pm 0.06$; Robot: $0.20 \pm 0.06$; Human: $0.21 \pm 0.09$. Kruskal-Wallis test, $\chi^2(2) = 0.55$, $p = 0.759$).

Interestingly, participants showed a remarkable distortion in their perception of own and partner's performance. Specifically, participants did not recognize that their partner was more accurate than they were, since their performance ratings were even higher for their own accuracy (Computer: $6.28 \pm 1.10$; Robot: $6.4 \pm 1.08$; Human: $6.16 \pm 1.43$) than that of the partner (Computer: $5.52 \pm 1.64$; Robot: $6.24 \pm 1.90$; Human: $5.52 \pm 1.48$). A mixed-effects model with performance rating as dependent variable, agent (self or other), identity of the other agent (computer, robot or human) as independent factors and subject as random effect confirms the presence of an effect of agent (higher ratings for self than other, omnibus test: $\chi^2(1) = 6.14$, $p = 0.013$) but no effect of partner's identity ($\chi^2(2) = 2.76$, $p = 0.251$) and interaction ($\chi^2(2) = 1.53$, $p = 0.466$).

### 6.2.2. Social influence under uncertainty

In line with the egocentric bias observed in participants' performance ratings, analysis of final decisions reveal that participants relied more on their own estimate than that of the partner (average influence, Computer: $0.26 \pm 0.13$; Robot: $0.35 \pm 0.16$; Human: $0.23 \pm 0.14$. Wilcoxon signed-rank test, null hypothesis: average influence = 0.5. Computer: $z = 4.29$, $r = 0.86$, $\eta^2 = 0.74$, $p < 0.001$; Robot: $z = 3.51$, $r = 0.70$, $\eta^2 = 0.49$, $p < 0.001$; Human: $z = 4.37$, $r = 0.87$, $\eta^2 = 0.76$, $p < 0.001$. Results are significant at the Bonferroni-corrected threshold for 3 comparisons). Although participants in all conditions showed egocentric advice discounting in their final decisions, their mean level of influence was statistically different across conditions (Kruskal-Wallis test, $\chi^2(2) = 7.67$, $p = 0.022$). We further explored this difference by running a mixed-effects model with trial-by-trial influence as dependent variable and experimental condition as categorical factor (Fig. 2a. Supplementary Information, model 1). Results reveal that participants were more influenced by the robot partner than both the human partner (Human – Robot, B = - 0.118, β = - 0.473, z = - 2.97, p = 0.003) and the computer one (Computer – Robot, B = - 0.084, β = - 0.338, z = - 2.12, p = 0.034). We did not find any difference between the Human and the Computer conditions (Human – Computer, B = - 0.034, β = - 0.135, z = - 0.85, p = 0.397). These results do not change when controlling for participants' estimation error (Supplementary Information, model 2). These differences are particularly interesting if we consider that the responses of the partner were identical across experimental conditions. This finding suggests that the nature of the partner (computer, robot or human) played a role in the processing and weighting of the partner's estimates, supporting H1.

---

[3] We acknowledge that part of the results of this set of analysis, for Robot and Computer conditions only, can be also found in previously published papers [60, 79]. We believe that reporting these results in the current section, together with results of the Human condition, is helpful for the interpretation of the results of the next paragraphs.

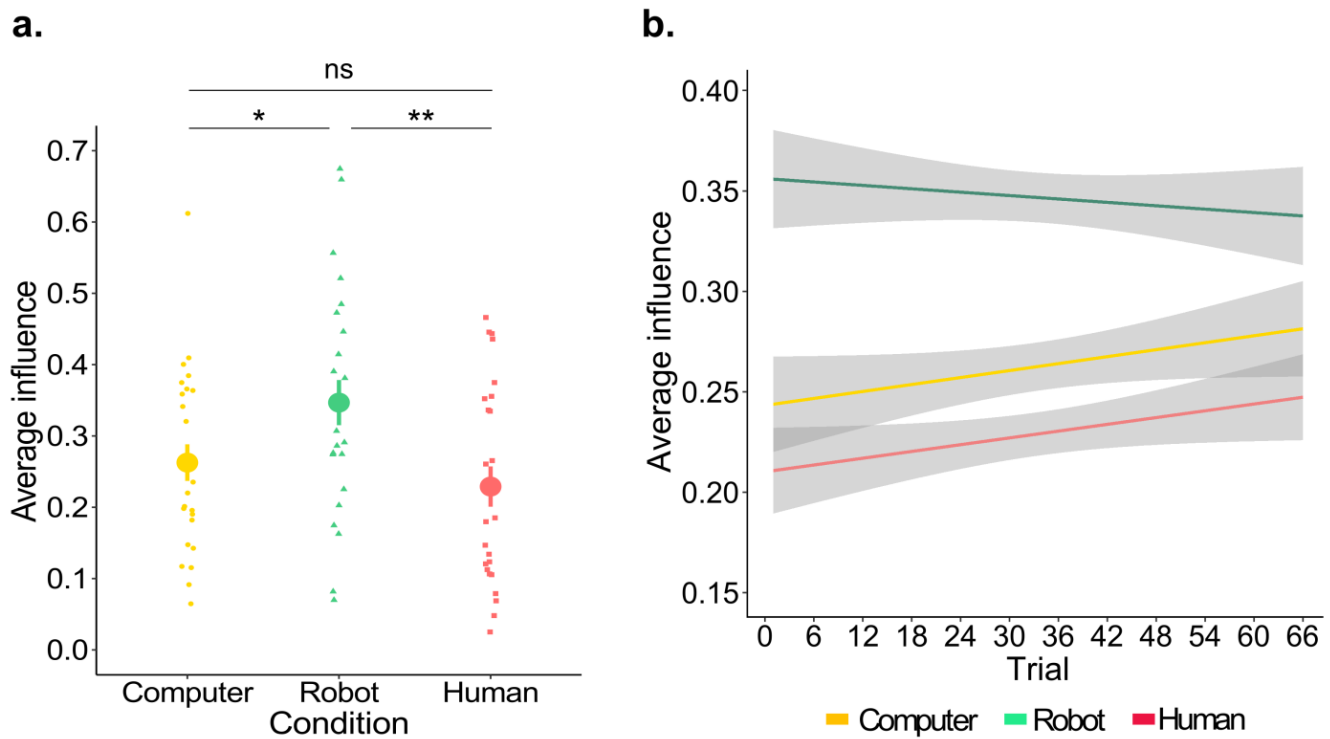

**Fig. 2. a)** Average influence in the Social influence task. Point-range plot of mean influence in Computer, Robot and Human conditions. Large dots represent mean values for each condition, bar ranges represent standard error of the mean and small symbols individual influence means. Influence was significantly different across conditions: mean influence was higher in the Robot condition than in the other two conditions. ** p < 0.01, * p < 0.05, ns = non-significant, mixed-effects model with condition as categorical factor. **b)** Temporal evolution of influence in the Social influence task. We characterize the temporal evolution of participants' influence along the 66 trials of the Social influence task. Data has been smoothed by linear fit separately for the three conditions, represented by different colors. Grey bounds represent standard errors.

**6.2.3. The impact of prior beliefs on social influence**

We further investigated the hypothesis that prior beliefs are crucial for the process of weighting of the partner's responses by analyzing the temporal evolution of participants' influence along the task (Fig. 2b). We ran a mixed-effects model with trial-by-trial influence as dependent variable and experimental condition, trial and their interaction as predictors (Supplementary Information, model 3). To facilitate the interpretation of the results of model 3, we also report B values obtained by running the same model with the influence variable as a percentage, instead of a proportion (B%, see Model 3 in the Supplementary information). One of the most interesting effects resulting from the current model is the difference between the intercepts across conditions. Indeed, the intercepts represent the estimated level of influence *at time zero* for each condition, which can be seen as an estimate of the participants' level of influence prior to temporal effects due to the repetitive exposure to the partner's feedback along the task. Results show that the estimated level of influence *at time zero* is significantly higher in the Robot condition than in Human and Computer conditions (Human – Robot, B = - 0.146, B% = -14.60%, $z = - 3.52$, $p < 0.001$; Computer – Robot, B = - 0.113, B% = -11.30%, $z = - 2.72$, $p = 0.007$), whereas we did not find any differences between Human and Computer conditions (Human – Computer, B = - 0.033, B% = -3.31%, $z = - 0.80$, $p = 0.426$). Then we investigated the presence of temporal effects in participants' influence along the task. In this case, the B% value represents, in form of a percentage, the average increase/decrease of influence *per trial*. Results show that participants in Human and Computer conditions significantly *increase* their level of influence along the task with very similar slopes (Computer: B = 0.001, B% = + 0.06%, $z = 2.23$, $p = 0.026$; Human: B = 0.001, B% = + 0.06%, $z = 2.17$, $p = 0.03$; interaction effect, Human - Computer, B = - 0.000, B% = - 0.00%, $z = - 0.04$, $p = 0.966$). The effect is absent in the Robot condition (B = - 0.000, B% = - 0.03%, $z = - 1.09$, $p = 0.276$), where the level of influence tends to *decrease* along time, converging towards the mean level of influence of participants in the other two conditions at the end of the experiment (Fig. 2b). Indeed, the interaction effects between the Robot condition and the other two conditions are significant (Human – Robot, B = 0.001, $z = 2.30$, $p = 0.021$; Computer – Robot, B = 0.001, $z = 2.35$, $p = 0.019$). Interestingly, if we multiply B% by the total number of trials, we can obtain an estimate of the general change in influence along the task in each of the three groups. If we combine this estimate with the intercept in the three conditions (expressing the level of influence at time "zero"), we can obtain a comprehensive estimate of the temporal evolution of influence in the three conditions. Indeed, participants in the Computer condition increase their level of influence from 24.32% to 28.14%; the Human group increases the level of influence from 21.02% to 24.73%, whereas participants in the Robot condition slightly decrease their level of influence from 35.62% to 33.76% along the experiment (Fig. 2b). These results suggest that the highest difference between the Robot condition and the other two conditions can be found at the very beginning of the experiment, when prior beliefs have a stronger influence on the interpretation of the partner's feedback. Conversely, repeated exposure to the partner's responses leads to a decrease in the discrepancy between participants' influence between the three conditions.

Furthermore, we tested whether the interaction between prior beliefs and the very first feedback useful for an estimate of the partner's competence could explain the participants' willingness to follow the partner's suggestions in the entire experiment, based on the experimental condition. We ran a linear regression with mean influence (averaged along the entire experiment) as dependent variable and condition, initial distance and their interactions as factors. Results reveal that large part of the difference between the three experimental conditions was due to the group of participants that observed a high distance in the first trial (initial distance = high). Indeed, participants in the Robot group interpreted this feedback as a signal that the robot should be taken into high consideration in their final decision, whereas participants in Human and Computer conditions adjusted their final decision to a much lower degree when they have seen a high discrepancy in the first trial of the task (Human – Robot, B = - 0.163, β = - 1.085, t = - 2.77, p = 0.007; Computer – Robot, B = - 0.150, β = - 0.998, t = - 2.55, p = 0.013). Conversely, if we consider the group of participants that observed a relatively lower distance in the first trial (initial distance = low), we do not observe statistically significant differences across conditions in terms of influence along the experiment (Human – Robot, B = - 0.076, β = - 0.510, t = - 1.35, p = 0.180; Computer – Robot, B = - 0.024, β = - 0.159, z = - 0.42, p = 0.674). Then we checked whether this result could be linked to a correlation between agents' distance in the first trial and mean agents' distance along the experiment. We ran the same regression by using the mean agents' distance (low or high) instead of initial distance as predictor. However, in this case we do not observe a significant difference across conditions in the "high distance" group (Human – Robot, B = - 0.103, β = - 0.690, t = - 1.76, p = 0.082; Computer – Robot, B = - 0.105, β = - 0.703, t = - 1.80, p = 0.077).

These results indicate that prior beliefs interacted in a peculiar way with the *very first* feedback on the partner's competence, playing a crucial role in the participants' interpretation of the partner's behavior and capabilities and participants' behavior along the task. In Human and Computer conditions, a high distance in the first trial convinced participants not to take into high consideration the partner's suggestion for the entire experiment; conversely, participants in the Robot group interpreted this remarkable response discrepancy as an indication that the robot's suggestions should have been taken into account for the optimization of task performance.

Altogether, these findings support H1, highlighting the crucial role of *a priori* beliefs in the interpretation of the partner's behavior and in the use (or disuse) of this information for upcoming decisions.

### 6.2.4. Partner's reliability, social influence and social norms

Then we investigated if within-subject modulation of influence was linked to the perceived reliability of the partner's estimates. Therefore, we tested the relationship between influence and agents' distance. We ran a mixed-effect linear regression with final decision as dependent variable, condition (Computer, Robot, Human), distance and their interactions as independent variables, with subject as random effect (Supplementary information, model 4). Results show a negative effect of distance on influence in all the three experimental conditions (Computer: B = - 0.435, β = - 0.291, z = - 12.86, p < 0.001; Robot: B = - 0.299, β = - 0.200, z = - 9.76, p < 0.001; Human: B = - 0.058, β = - 0.039, z = - 1.98, p = 0.048) groups: more specifically, participants' influence decreased with the increase of the distance from the partner's estimate (Figure 3b).

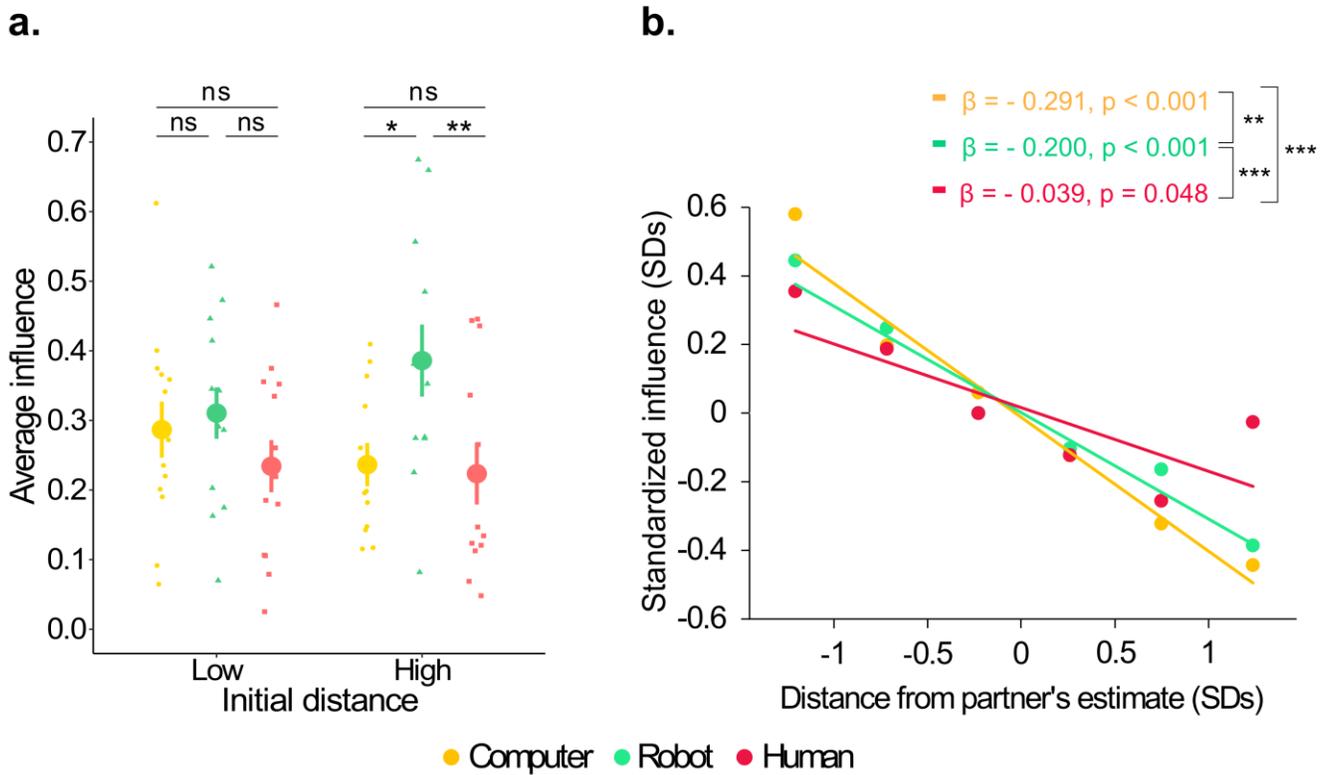

**Fig 3. a)** Influence as a function of agents' distance in the first trial. Point-range plot of mean influence (along the entire experiment) based on agents' distance in the first trial (initial distance: high or low) in Computer, Robot and Human conditions. Large dots represent average values for each level of initial distance by condition, bar ranges represent standard error of the mean and small symbols individual influence means. ** p < 0.01, * p < 0.05, ns = non-significant, multiple regression with initial distance, condition and their interactions as factors. **b)** Influence as a function of agents' distance. Standardized (within-subject) influence plotted as a function of distance from the partner's estimate. To obtain this graph, for each participant, we standardized trial-by-trial normalized distance between the estimates of the two agents (distance (cm) / stimulus length (cm)) to express each distance value in terms of deviation from the individual subject mean. We clustered standardized distances in six bins to obtain six distance ranges with a similar mean number of observations per bin: 1) distance < -1 SD; 2) -1 SD ≥ distance < -0.5 SD); 3) –0.5 SD ≥ distance < 0 SD; 4) 0 SD ≥ distance < 0.5 SD; 5) 0.5 SD ≥ distance < 1 SD; 6) distance ≥ 1 SD. We also standardized, for each participant, trial-by-trial influence to express influence values in terms of deviation from individual influence means. Eventually, for each distance range, we averaged individual influence means across subjects. We report standardized regression coefficients and p-values obtained through a mixed-effect model analyzing the relationship between within-subject modulation of influence by agents' distance. We also report interactions effects assessing between-condition differences in the relationship between influence and response distance: ** p < 0.01, *** p < 0.001.

One the one hand, this effect suggest that participants' susceptibility to their partner was indeed modulated by the perceived reliability of the current partner's feedback; one the other hand, it reveals that the existence of a discrepancy between the two interacting partners' responses was interpreted by participants as a signal of the unreliability of the partner rather than themselves, confirming the emergence of egocentric advice discounting. Moreover, we highlight the existence of interaction effects: the effect of distance was stronger in the Computer condition than in the other two conditions (Robot - Computer, B = 0.136, β = 0.091, z = 2.99, p = 0.003; Human - Computer, B = 0.377, β = 0.252, z = 8.40, p < 0.001). Moreover, the effect of distance was stronger in the Robot than in the Human condition (Human - Robot, B = 0.241, β = 0.161, z = 5.66, p < 0.001). Altogether, interaction effects suggest that the nature of the partner determines the extent to which participants' use its perceived reliability to modulate their expression of susceptibility to them. In particular, these results suggest that the more the partner shows human-like or social characteristics, the less participants' susceptibility depends on the perceived reliability of the partner's estimates. We further explored this hypothesis by running a regression with influence as dependent variable and condition, performance ratings (self – other) and their interactions as independent variables. Results reveal significant interactions highlighting a stronger relationship between influence and performance ratings in Computer and Robot conditions compared to the Human condition (Robot - Human, B = - 0.054, β = - 0.667, t = - 2.34, p = 0.022; Computer - Human, B = - 0.047, β = - 0.581, t = - 2.05, p = 0.044. Figure 4). We did not find a significant difference between Robot and Computer conditions (Robot – Computer, B = - 0.007, β = - 0.085, t = - 0.38, p = 0.705). These interaction effects were driven by a significant main effect in Robot (B = - 0.043, β = - 0.531, t = - 3.32, p = 0.001) and Computer (B = - 0.036, β = - 0.446, t = - 2.83, p = 0.006) conditions, which was absent in the Human condition (B = 0.011, β = 0.136, t = 0.58, p = 0.567).

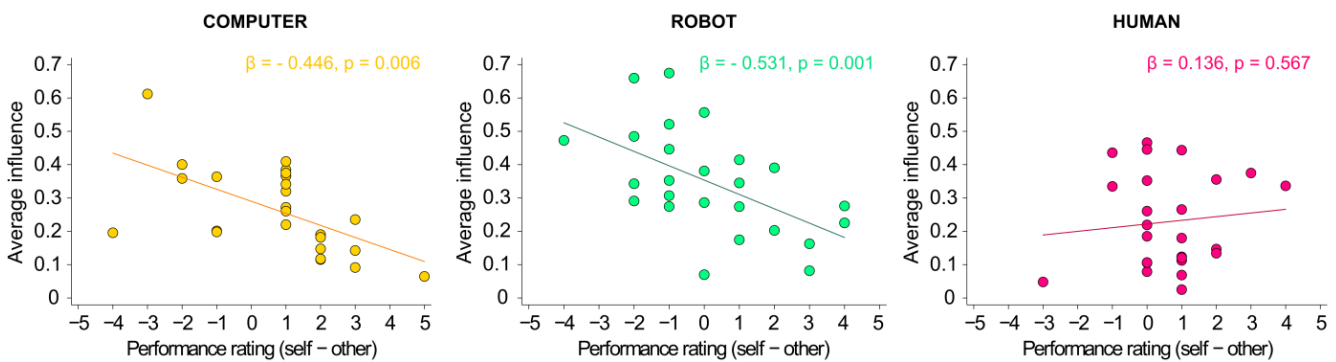

**Fig. 4.** Scatter plots of individual average influence as a function of the difference between self and partner rating of estimation accuracy, assessed at the end of the Social influence task. We found a positive relationship between performance ratings and mean influence in Computer and Robot conditions, whereas there is no significant relationship in the Human condition. Standardized beta values (β) and p-values are derived from the main effects of regression with mean influence as dependent variable and condition as categorical factor.

These findings underline a marked difference between participants' willingness to rely on the judgment of either a mechanical agent or a peer. One the one hand, reliance on the judgments of a mechanical partner (computer or robot) was strictly dependent on the perceived reliability of the mechanical agent itself. On the other hand, the explicit expression of susceptibility towards a human partner was not fully explained by the perceived relative ability of the two interacting partners, suggesting the emergence of social normative dynamics in peer interaction. These results support H2, highlighting important differences in the emergence of social norms during interaction with human peers compared to non-human agents.

## 7. Discussion

### 7.1 The role of prior beliefs in social influence under uncertainty

When interacting with other agents, we may or may not rely on their advice based on the perception of our own and their competence and performance in the current task. This is the case for interaction with human peers, but also for interaction with humanoid social robots or artificial agents [33, 97]. However, in interactive settings we do not always have access to objective and accurate information about individuals' performance and abilities. In this context, our prior beliefs on the capabilities and relational properties of other agents can bias our interpretation of their behavior and our willingness to rely on information provided by them.

In the current study, participants performed a joint perceptual task in three between-subject conditions (Human, Robot and Computer): participants in the three groups believed to perform the task with different partners (a human participant, a humanoid social robot iCub or a computer, respectively). Participants had to estimate the length of visual stimuli and, after each decision, could observe the estimate of their respective partner. In fact, the estimates of the interacting partners were systematically controlled by the same algorithm in all the conditions. After feedback on the partner's judgment, participants had the possibility to modify their own response (final decision), revealing the influence exerted by the partner on them.

Results reveal a significant effect of condition in the level of susceptibility to the partner's advice in participants' final decisions. In particular, participants in the Robot condition were more influenced by the partner's advice than those in the other two conditions, although the actual accuracy of the interacting partners was comparable across conditions. This difference was particularly high at the beginning of the experiment, whereas it tended to decrease along the experiment after repetitive feedback about the partners' responses. These findings highlight the importance of prior beliefs about the nature and the capabilities of other agents (human, robotic or algorithmic) in determining reliance on them [98]. This is especially true in all the contexts in which individuals cannot interact repetitively for a sufficient period of time and cannot receive relevant feedback signaling the competence of the interacting partner(s). Several factors, in principle, might have determined this result in our experimental paradigm. The first and more interesting question concerns the drivers of the difference between the Robot and the Human condition, which might involve social influence mechanisms related to 1) attribution of competence to the partner and 2) social normative mechanisms modulating the overt expression of susceptibility towards the partner.

**7.2. Human vs. robot partner: prior beliefs and perception of competence**

The first dimension concerns the impact of prior beliefs on the perception of the accuracy and the reliability of the perceptual and motor systems sustaining the partner's response. In this regard, we highlight that participants in the Robot condition had a brief experience of the existence of a perceptual system in the humanoid robot iCub before starting the experiment. In particular, through a brief meeting with iCub, participants were implicitly informed about the task-related capabilities of the robot. Participants had the impression that iCub was capable of making autonomous and accurate movements in order to reach and touch a touch-screen placed in front of it. Moreover, participants understood that iCub was endowed with visual perception and was indeed able to see the visual stimuli shown on the tablet. We stress that, in this preliminary encounter with the robot, participants did *not* receive any direct or indirect clue about the actual accuracy of the robot. Indeed, we believe that the higher levels of susceptibility observed in the Robot condition are grounded in the relatively high "a priori" confidence that humans genuinely attribute to robotic systems [27], which can be maintained in absence of feedback revealing failures or inefficient behavior [65]. Concerning the Human condition, it is likely that participants assumed that, in principle, they and their human partner possessed very similar perceptual skills. However, the observed (meta-cognitive) inability to assess own and other's performance in absence of accuracy feedback, reinforced by the typical tendency to over-estimate own capabilities in social contexts [5 52], led to low confidence in the perceptual abilities of the partner. These results are also consistent with evidence in social decision-making revealing that humans' susceptibility to information provided by others depends on the relative confidence of the interacting agents and on the possibility to communicate accurately their actual level of confidence to each other [41, 99, 100].

Interestingly, the difference between Human and Robot conditions was mainly driven by the interaction between prior beliefs and initial signals about the partner's competence. Participants who had observed a relatively higher discrepancy between their own and their partner's responses, at the very beginning of the experiment, behaved differently depending on the experimental condition. Participants in the Robot condition likely interpreted this high discrepancy as an indication of their own fallibility in the task and decided to follow the advice of the robot to a relatively higher extent; conversely, participants in the Human condition interpreted this high discrepancy as a signal of the unreliability of the partner, leading to lower levels of influence. This findings indicate that prior beliefs on other's competence may have a strong impact on the interpretation of minimal and uncertain piece of evidence linked to others' reliability. We believe these findings to have important implications for research of trust in HRI. This high "a priori" confidence in the robot's ability, which have an impact on repetitive interaction with the robot itself, can represent a double-edged sword for human-robot interaction. When, as in our experiment, the robot produces reliable and effective behavior, this high confidence in the robot could help the human partner in trusting the robot during cooperative interaction, with positive implication for assistive real-life scenarios (see the paragraph "7.5. Implications"). However, the observed high "a priori" trust in the robot competence could be detrimental in the presence of faulty and unreliable robots, in line with several findings showing over-trust in robots [28-31].

## 7.3. Human vs. robot partner: social norms and social influence

The second dimension that could have an impact on participants' influence concerns their willingness to overtly reveal their actual level of confidence in the partner's competence. In this regard, we remind that, in the Social influence task, participants were told that their partner would observe their final decisions. Extensive evidence in the human-human interaction literature has demonstrated that the overt expression of opinions and judgments in social contexts is modulated by normative conformity and peer pressure [6-8, 56-61, 101], which distort the process of weighting of information provided by others depending on its actual reliability, which is predicted by Bayesian theories of information aggregation [58, 102]. This interpretation is corroborated by our results that show no between-subject relationship between performance ratings (self – other) and influence in the Human condition; conversely we found a strong relationship between performance ratings and participants' susceptibility in Computer and Robot conditions. Moreover, within-subject modulation of trial-by-trial influence based on the current distance from the partner's response was much less pronounced in the Human condition than in the other two experimental conditions. These findings suggest that participants' belief of interacting with another human participant interfered in a peculiar way with the process of weighting of information by reliability. In particular, participants might have decided to adjust their final estimates by a fair amount even when participants believed the partner's response to be unreliable (i.e., when their responses were inconsistent between each other), following normative principles. Conversely, in Robot and Computer conditions participants' reliance on the partner rapidly and monotonically decreased as long as the distance from the partner increased, highlighting a more systematic decision process based on the estimation of the partner's accuracy in the current trial. This is consistent with extensive evidence revealing that social influence in human-robot and human-machine scenarios is primarily shaped by the perception of the agent's competence and performance [22-26].

An interesting question is whether normative consideration could have played a role also in the Robot condition. Indeed, the human-like and social characteristics of the humanoid robot iCub might have led participants to try to please the robot by showing (relatively) higher levels of susceptibility. This hypothesis is motivated by studies showing that humanoid robots can evoke automatic behavioral reactions similar to those exerted by humans [103]. Actions of both humanoid robots and humans trigger, in an observer, similar responses in terms of motor resonance [104, 105], anticipation [90] and speed adaptation [106] in simple action observation tasks. Most importantly, extensive evidence has revealed the emergence of pro-social behavior towards robots in adults [66, 68, 69, 72, 107] and children [108-111]. However, evidence revealing the emergence of purely normative conformity in mixed (human-robot) dyads or groups is still inconclusive. Several studies failed in revealing an effect of normative conformity in adults while interacting with robots [80-82, 85], although mechanisms of normative conformity have been found in adult dyads [79] children and adults who interact with a group of robots [80-83] and in groups of one robot and several human adults [84]. Moreover, recent evidence has shown that humans' willingness to trust or imitate robots' behavior is tightly related with the uncertainty of the task solution [85]. Furthermore, in our task, the robot did not provide socially relevant signals to the participant, since the only feedback concerning the behavior of the robot consisted in its perceptual estimates, which had purely functional value. In this context, it is unlikely that participants used social normative considerations when making their final decisions, in line with the idea that the emergence of affective and pro-social behavior towards robots requires explicit and transparent social cues from the robot itself [79].

## 7.4. Social influence in interaction with a computer algorithm

Concerning the Computer condition, it is unlikely that mechanisms related to the attribution of perceptual–motor competence and social norms played a role in the modulation of participants' social influence. In fact, the computer partner did not possess any kind of perceptual or motor system. We believe that participants' influence in the Computer condition depended entirely on the estimated performance of the algorithm that produced the perceptual estimates, in line with mechanisms of informational conformity [112], and therefore was not influenced by social normative considerations, as generally observed in interaction with fully autonomous virtual agents [113]. In this regard, since participants decreased their level of susceptibility as the distance from the partner's estimate increased, we hypothesize that participants believed that the partner's accuracy has been varying according to a certain response function. Due to the observed scarce self-assessment of own and partner's accuracy and consequent egocentric bias, the general average level of influence in the Computer condition was rather low and, interestingly, significantly lower than that the one observed in the Robot condition.

## 7.5. Implications

Our results show that humans' *a priori* beliefs on robots' competence could play a crucial role in their willingness to accept help and advice from them. In our experimental task, these beliefs modulate the perception, the evaluation and the use of the robot's advice under uncertainty. In particular, our findings reveal that the absence of feedback on the robot's performance does not prevent and even *increase* individuals' probability to use the partner's advice compared to interaction with a human peer or a computer algorithm. This phenomenon might play a role in real-life contexts involving assistive and service robots. In these scenarios, we need robots that can be trusted by customers, workers or patients that may benefit from the robot's assistance. Specifically, our study focuses on those situations in which individuals cannot learn the reliability of the robot from direct experience (i.e., through feedback on its performance), but rather should "trust" the advice-giving robot in the presence of uncertainty on its competence. For instance, our findings can have implications on service robots welcoming and assisting customers in stores, hotels, restaurants, banks and other commercial activities, or providing assistance in hospitals, factories and schools. Service providers using robots in their activities should be aware that, independently of the appearance or behavior of robots themselves, the evaluation of their competence is strongly biased by the customers' prior beliefs on the robot's competence in the current task. These beliefs may depend on several factors, including: previous experience with robots, general trust in technology, risk-aversion and, most of all, the type of task in which the robot is involved. Indeed, the type of service provided by the robot and the relative context is crucial in determining the individuals' level of a priori trust towards the robot itself, which may be high in certain types of task, but low in others [77, 78]. Furthermore, our findings show that prolonged access to the robot's behavior leads to a decrease in the impact of prior beliefs, even in the absence of precise feedback on the robot's performance. Humans naturally estimate the robot's competence observing the robot's behavior: if they have reasons to believe that the robot is doing mistakes, they will likely and rapidly lose trust in its competence and will stop to follow its advice [63-65]. This aspect should be always taken into account by entities providing services consisting of repeated interactions with the same robot, including banks, hospitals and companies designing robots for domestic assistance.

## 8. Limitations

The present study investigated the role of prior beliefs in social influence under uncertainty in the context of a functional perceptual task. Recent evidence has highlighted marked differences between functional and social tasks concerning human willingness to rely on robots [77-78]. Therefore, the results of our paper (e.g., higher influence in the Robot condition) may change in tasks targeting "social" abilities (e.g., theory of mind, moral judgment, empathy), where humans may be expected to exhibit more refined skills than robots. For this reason, researchers should be careful in generalizing our results to other kinds of interactive contexts, especially those involving social skills.

Furthermore, we acknowledge that the current experimental task did not require any physical interaction with the robot. Furthermore, the robot did not produce any kind of socially relevant behavior during the task. In this context, it is not surprising that our results did not provide clear evidence on the emergence of social normative mechanisms modulating social influence under uncertainty in human-robot interaction. Although the peculiar features of our experiment guaranteed the comparability of our experimental conditions on a functional level, they require caution concerning the generalizability of our results to scenarios involving physical interaction between humans and robots, especially if they entail affective processes. Indeed, extensive findings in HRI [e.g., 67, 68, 73, 114] have revealed the emergence of emotional and empathic reactions in human participants during interaction with a physically present robot, particularly if endowed with social-like and human-like behavior. Researchers should be aware of the growing literature on the social, affective and normative effects modulating human-robot interaction if they aim at comparing our findings to other human-robot interaction scenarios.

# 9. Conclusions and future directions

The current study aimed at exploring the differences and the similarities underlying social influence under uncertainty in interaction with peers, social robots and computers, using a unique and controllable experimental protocol. Our results suggest that, in the absence of feedback about the interacting partner's performance, the influence exerted by the partner is biased by participants' prior beliefs about its nature (human, robot or computer) and relative competence, leading to distortions in the process of weighting of socially-acquired information. Moreover, we show that human-human social influence is characterized by the emergence of social normative mechanisms that interfere with a process of integration of information that is uniquely based on its reliability. These normative phenomena do not appear to emerge in human-robot social influence, which is modulated by informational mechanisms similar to those intervening in human-computer interaction.

Our findings offer novel insights in the understanding of the informational and social normative mechanisms underlying humans' susceptibility to peers and autonomous agents. We hope that our work could fuel further research on social influence under uncertainty in human-robot interaction. Although our experimental task specifically targeted perceptual abilities, human-robot interaction is a much richer process requiring other types of skills (e.g., motor, cognitive, social). Future studies may employ different tasks and interactive scenarios to understand the role of prior beliefs in other types of robots' behavior, pointing in the direction of a more integrated and exhaustive view of human-robot social influence. For instance, future works may investigate the impact of social and affective signals produced by social robots on social influence under uncertainty.

Future studies may also try to control and manipulate task difficulty and participants' uncertainty in order to understand the circumstances under which individuals may be more willing to rely on robotic partners. Furthermore, future studies may enrich the investigation of human-robot interaction by allowing robots themselves to express their own level of confidence in the human partner's abilities [79, 115], in line with recent cognitive architectures modelling human-robot trust-based behavior from a robot-centered perspective [116, 117]. These dynamics may reveal crucial in the design of robotic agents that could effectively act as collaborative companions in contexts such as healthcare [16, 18], rehabilitation [118], elderly people assistance [17], customer care [19, 119] and education [20-21].

46. Sniezek, J. A., Schrah, G. E., & Dalal, R. S. (2004). Improving judgement with prepaid expert advice. *Journal of Behavioral Decision Making, 17*(3), 173-190. 10.1002/bdm.468

47. Gardner, P. H., & Berry, D. C. (1995). The effect of different forms of advice on the control of a simulated complex system. *Applied Cognitive Psychology*, *9*(7), S55–S79. 10.1002/acp.2350090706

48. Toelch, U., Bach, D. R., & Dolan, R. J. (2014). The neural underpinnings of an optimal exploitation of social information under uncertainty. *Social Cognitive and Affective Neuroscience*, *9*(11), 1746-1753. 10.1093/scan/nst173

49. Yaniv, I., & Kleinberger, E. (2000). Advice taking in decision making: Egocentric discounting and reputation formation. *Organizational Behavior and Human Decision Processes*, *83*(2), 260-281. 10.1006/obhd.2000.2909

50. Heyes, C. (2012). What's social about social learning?. *Journal of Comparative Psychology*, *126*(2), 193-202. 10.1037/a0025180

51. Soll, J. B., & Larrick, R. P. (2009). Strategies for revising judgment: How (and how well) people use others' opinions. *Journal of experimental psychology: Learning, memory, and cognition*, *35*(3), 780–805. 10.1037/a0015145

52. Duval, T. S., & Silvia, P. J. (2002). Self-awareness, probability of improvement, and the self-serving bias. *Journal of personality and social psychology*, *82*(1), 49-61. 10.1037/0022-3514.82.1.49

53. Molleman, L., Tump, A. N., Gradassi, A., Herzog, S., Jayles, B., Kurvers, R. H., & van den Bos, W. (2020). Strategies for integrating disparate social information. *Proceedings of the Royal Society B, 287*(1939), 20202413. 10.1098/rspb.2020.2413

54. Mahmoodi, A., Bang, D., Olsen, K., Zhao, Y. A., Shi, Z., Broberg, K., et al.. (2015). Equality bias impairs collective decision-making across cultures. *Proceedings of the National Academy of Sciences, 112*(12), 3835-3840. 10.1073/pnas.142169211

55. Krueger, X. (2003). Return of the ego–self-referent information as a filter for social prediction: comment on Karniol (2003). *Psychological Review, 110*, 585–590; discussion 10.1037/0033-295x. 110.3.585

56. Festinger, L. (1957). A theory of cognitive dissonance (Vol. 2). Stanford university press.
31